# Evaluation Evaluation
# a Monte Carlo study

D.M.W. Powers[1]

**Abstract.** Over the last decade there has been increasing concern about the biases embodied in traditional evaluation methods for Natural Language Processing/Learning, particularly methods borrowed from Information Retrieval. Without knowledge of the Bias and Prevalence of the contingency being tested, or equivalently the expectation due to chance, the simple conditional probabilities Recall, Precision and Accuracy are not meaningful as evaluation measures, either individually or in combinations such as F-factor.

The existence of bias in NLP measures leads to the 'improvement' of systems by increasing their bias, such as the practice of improving tagging and parsing scores by using most common value (e.g. water is always a Noun) rather than the attempting to discover the correct one. The measures Cohen Kappa and Powers Informedness are discussed as unbiased alternative to Recall and related to the psychologically significant measure DeltaP.

In this paper we will analyze both biased and unbiased measures theoretically, characterizing the precise relationship between all these measures as well as evaluating the evaluation measures themselves empirically using a Monte Carlo simulation.

## 1  INTRODUCTION

A common but poorly motivated way of evaluating results of Language and Learning experiments is using Recall, Precision and F-factor. These measures are named for their origin in Information Retrieval and present specific biases, namely that they ignore performance in correctly handling negative examples, they propagate the underlying marginal Prevalences and Biases, and they fail to take account the chance level performance. In the Medical Sciences, Receiver Operating Characteristics (ROC) analysis has been borrowed from Signal Processing to become a standard for evaluation and standard setting, comparing True Positive Rate and False Positive Rate. In the Behavioural Sciences, Specificity and Sensitivity, are commonly used. Alternate techniques, such as Rand Accuracy, have some advantages but are nonetheless still biased measures unless explicitly debiased. We will recapitulate some of the literature relating to the problems with these measures, as well as considering a number of other techniques that have been introduced and argued within each of these fields, aiming/claiming to address the problems with these simplistic measures.

This paper examines the relationships between these various measures, develops new insights into the problem of measuring the effectiveness of an empirical decision system or a scientific experiment, analyzing and introducing new probabilistic and information theoretic measures that overcome the problems with Recall, Precision and their derivatives.

[1] AILab, CSEM, Flinders University of South Australia, email:David.Powers@flinders.edu.au

## 2  THE BINARY CASE

It is common to introduce the various measures in the context of a dichotomous binary classification problem, where the labels are by convention + and − and the predictions of a classifier are summarized in a four cell contingency table. This contingency table may be expressed using raw counts of the number of times each predicted label is associated with each real class, or may be expressed in relative terms. Cell and margin labels may be formal probability expressions, may derive cell expressions from margin labels or vice-versa, may use alphabetic constant labels `a, b, c, d` or `A, B, C, D,` or may use acronyms for the generic terms for True and False, Real and Predicted Positives and Negatives. `UPPER CASE` typewriter font is used where the values are counts, and `lower case` where the values are probabilities or proportions relative to `N` or the marginal probabilities; in addition will use Mixed Case text font for popular nomenclature that may or may not correspond directly to one of our formal systematic names. True and False Positives (`TP/FP`) refer to the number of Predicted Positives that were correct/incorrect, and similarly for True and False Negatives (`TN/FN`), and these four cells sum to `N`. On the other hand `tp, fp, fn, tn` and `rp, rn` and `pp, pn` refer to the joint and marginal probabilities, and the four contingency cells and the two pairs of marginal probabilities each sum to 1. We will attach other popular names to some of these probabilities in due course.

We thus make the specific assumptions that we are predicting and assessing a single condition that is either positive or negative (dichotomous), that we have one predicting model, and one gold standard labelling. Unless otherwise noted we will also for simplicity assume that the contingency is non-trivial in the sense that both positive and negative states of both predicted and real conditions occur, so that no marginal sums or probabilities are zero.

We illustrate in Table 1 the general form of a binary contingency table using both the traditional alphabetic notation and the directly interpretable systematic approach. Both definitions and derivations in this paper are made relative to these labellings, although English terms (e.g. from Information Retrieval) will also be introduced for various ratios and probabilities. The positive diagonal represents correct predictions, and the negative diagonal incorrect predictions. The predictions of the contingency table may be the predictions of a theory or grammar, of some computational rule or system (e.g. an Expert System or a Neural Network or a POS Tagger), or may simply be a direct measurement, a calculated metric, or a latent condition, symptom or marker. We will refer generically to "the model" as the source of the predicted labels, and "the population" or "the world" as the source of the real conditions. We are interested in understanding to what extent the model "informs" predictions about the world/population, and the world/population "marks" conditions in the model.

**Table 1.** Systematic and traditional notations in a contingency table.

|    | +R | −R |    |     | +R  | −R  |     |
|----|----|----|----|-----|-----|-----|-----|
| +P | tp | fp | pp | +P  | A   | B   | A+B |
| −P | fn | tn | pn | −P  | C   | D   | C+D |
|    | rp | rn | 1  |     | A+C | B+D | N   |

## 2.1 Recall & Precision, Sensitivity & Specificity

Recall or Sensitivity (as it is called in Psychology) is the proportion of Real Positive cases that are correctly Predicted Positive. This measures the Coverage of the Real Positive cases by the **+P** (Predicted Positive) rule. Its desirable feature is that it reflects how many of the relevant cases the **+P** rule picks up. It tends not to be very highly valued in Information Retrieval (on the assumptions that there are many relevant documents, that it doesn't really matter which subset we find, that we can't know anything about the relevance of documents that aren't returned). Recall tends to be neglected or averaged away in Machine Learning and Computational Linguistics (where the focus is on how confident we can be in the rule or classifier). However, in a Computational Linguistics/Machine Translation context Recall has been shown to have a major weight in predicting the success of Word Alignment [1]. In a Medical context Recall is moreover regarded as primary, as the aim is to identify all Real Positive cases, and it is also one of the legs on which ROC analysis stands. In this context it is referred to as True Positive Rate (`tpr`). Recall is defined, with its various common appellations, by equation (1):

$$
\begin{aligned}
\text{Recall} \quad &= \text{Sensitivity} = \texttt{tpr} = \texttt{tp/rp} \\
&= \texttt{TP / RP} = \texttt{A /(A+C)} \quad (1)
\end{aligned}
$$

Conversely, Precision or Confidence (as it is called in Data Mining) denotes the proportion of Predicted Positive cases that are correctly Real Positives. This is what Machine Learning, Data Mining and Information Retrieval focus on, but it is totally ignored in ROC analysis. It can however analogously be called True Positive Accuracy (`tpa`), being a measure of accuracy of Predicted Positives in contrast with the rate of discovery of Real Positives (`tpr`). Precision is defined in (2):

$$
\begin{aligned}
\text{Precision} \quad &= \text{Confidence} = \texttt{tpa} = \texttt{tp/pp} \\
&= \texttt{TP / PP} = \texttt{A /(A+B)} \quad (2)
\end{aligned}
$$

These two measures and their combinations focus only on the positive examples and predictions, although between them they capture some information about the rates and kinds of errors made. However, neither of them captures any information about how well the model handles negative cases. Recall relates only to the **+R** column and Precision only to the **+P** row. Neither of these takes into account the number of True Negatives. This also applies to their Arithmetic, Geometric and Harmonic Means: A, G and F=G2/A (the F-factor or F-measure). Note that the F-measure effectively references the True Positives to the Arithmetic Mean of Predicted Positives and Real Positives, being a constructed rate normalized to an idealized value. The Geometric Mean of Recall and Precision (G-measure) effectively normalizes TP to the Geometric Mean of Predicted Positives and Real Positives, and its Information content corresponds to the Arithmetic Mean of the Information represented by Recall and Precision.

In fact, there is in principle nothing special about the Positive case, and we can define Inverse statistics in terms of the Inverse problem in which we interchange positive and negative and are predicting the opposite case. Inverse Recall or Specificity is thus the proportion of Real Negative cases that are correctly Predicted Negative (3), and is also known as the True Negative Rate (`tnr`). Conversely, Inverse Precision is the proportion of Predicted Negative cases that are indeed Real Negatives (4), and can also be called True Negative Accuracy (`tna`):

$$
\begin{aligned}
\text{Inverse Recall} \quad &= \texttt{tnr} = \texttt{tn/rn} \\
&= \texttt{TN/RN} = \texttt{D/(B+D)} \quad (3) \\
\text{Inverse Precision} \quad &= \texttt{tna} = \texttt{tn/pn} \\
&= \texttt{TN/PN} = \texttt{D/(C+D)} \quad (4)
\end{aligned}
$$

Rand Accuracy explicitly takes into account the classification of negatives, and is expressible (5) both as a weighted average of Precision and Inverse Precision and as a weighted average of Recall and Inverse Recall. Conversely, the Jaccard or Tanimoto similarity coefficient explicitly ignores correctly classificatied negatives (TN):

$$
\begin{aligned}
\text{Accuracy} \quad &= \texttt{tea = ter = tp+tn} \\
&= \texttt{rp*tpr+rn*tnr} \\
&= \texttt{pp*tpa+pn*tna = (A+D)/N} \quad (5) \\
\text{Jaccard} \quad &= \texttt{tp/(tp+fn+fp) = TP/(N−TN)} \\
&= \texttt{A/(A+B+C)} = \texttt{A/(N−D)} \quad (6)
\end{aligned}
$$

Each of the above also has a complementary form defining an error rate, of which some have specific names and importance: Fallout or False Positive Rate (`fpr`) are the proportion of Real Negatives that occur as Predicted Positive (ring-ins); Miss Rate or False Negative Rate (`fnr`) are the proportion of Real Positives that are Predicted Negatives (false-drops). False Positive Rate is the second of the legs on which ROC analysis is based.

$$
\begin{aligned}
\text{Fallout} \quad &= \texttt{fpr} = \texttt{fp/rp} \\
&= \texttt{FP/RP} = \texttt{B/(B+D)} \quad (7) \\
\text{Miss Rate} \quad &= \texttt{fnr} = \texttt{fn/rn} \\
&= \texttt{FN/RN} = \texttt{C/(A+C)} \quad (8)
\end{aligned}
$$

Note that FN and FP are sometimes referred to as Type I and Type II Errors, and the rates `fn` and `fp` as alpha and beta, respectively – referring to falsely rejecting or accepting a hypothesis. More correctly, these terms apply specifically to the meta-level problem of whether the precise pattern of counts (not rates) in the contingency table fit the null hypothesis of random distribution rather than reflecting the effect of some alternative hypothesis (which is not in general the one represented by either **+P –> +R** or **−P –> −R** or both).

## 2.2 Prevalence, Bias, Cost & Skew

We now turn our attention to various forms of bias that detract from the utility of all of the above surface measures [2]. We will first note that `rp` represents the Prevalence of positive cases, RP/N, and is assumed to be a property of the population of interest – it may be constant, or it may vary across subpopulations, but is regarded here as not being under the control of the experimenter. By contrast, `pp` represents the (label) Bias of the model [3], the tendency of the model to output positive labels, `PP/N`, and is directly under the control of the experimenter, who can change the model by changing the theory or algorithm, or some parameter or threshold, to better fit the world/population being modelled. Note that F-factor effectively references `tp` (probability or proportion of True Positives) to the Arithmetic Mean of Bias and Prevalence. A common rule of thumb, or even a characteristic of some algorithms, is to parameterize a model so that Prevalence = Bias, viz. `rp = pp`. Corollaries of this setting are Recall = Precision (= A = G = F), Inverse Recall = Inverse Precision and Fallout = Miss Rate.

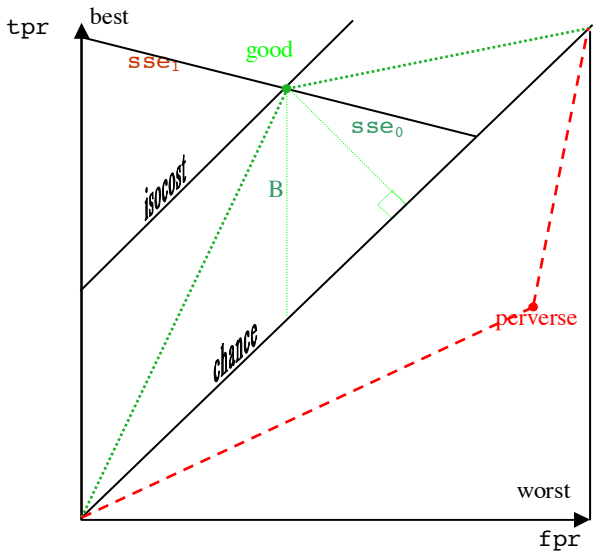

**Figure 1.** Illustration of ROC Analysis. The main diagonal represents chance with parallel isocost lines representing equal cost-performance. Points above the diagonal represent performance better than chance, those below worse than chance.

Alternate characterizations of Prevalence are in terms of Odds [4] or Skew [5], being the Class Ratio $c_s$ = `rn/rp`, recalling that by definition `rp+rn = 1` and `RN+RP = N`. If the distribution is highly skewed, typically there are many more negative cases than positive, this means the number of errors due to poor Inverse Recall will be much greater than the number of errors due to poor Recall. Given the cost of both False Positives and False Negatives is equal, individually, the overall component of the total cost due to False Positives (as Negatives) will be much greater at any significant level of chance performance, due to the higher Prevalence of Real Negatives.

Note that the normalized binary contingency table with unspecified margins has three degrees of freedom – setting any three non–Redundant ratios determines the rest (setting any count supplies the remaining information to recover the original table of counts with its four degrees of freedom). In particular, Recall, Inverse Recall and Prevalence, or equivalently tpr, fpr and $c_s$, suffice to determine all ratios and measures derivable from the normalized contingency table, but `N` is also required to determine significance. As another case of specific interest, Precision, Inverse Precision and Bias, in combination, suffice to determine all ratios or measures.

We can also take into account a differential value for positives (`cp`) and negatives (`cn`) – this can be applied to errors as a cost (loss or debit) and/or to correct cases as a gain (profit or credit), and can be combined into a single Cost Ratio $c_v$ = `cn/cp`. Note that the value and skew determined costs have similar effects, and may be multiplied to produce a single skew-like cost factor $c$ = $c_v c_s$. Formulations of measures that are expressed using tpr, fpr and $c_s$ may be made cost-sensitive by using $c$ = $c_v c_s$ in place of $c$ = $c_s$, or can be made skew/cost-insensitive by using $c$ = 1 [5].

### 2.3 ROC and PN Analyses

Flach [5] has highlighted the utility of ROC analysis to the Machine Learning community, and characterized the skew sensitivity of many measures in that context, utilizing the ROC format to give geometric insights into the nature of the measures and their sensitivity to skew. ROC analysis plots the rate `tpr` against the rate `fpr`. A perfect classifier will score in the top left hand corner (`fpr=0,tpr=100%`). A worst case classifier will score in the bottom right hand corner (`fpr=100%,tpr=0`). A random classifier would be expected to score somewhere along the positive diagonal (`tpr=fpr`) since the model will throw up positive and negative examples at the same rate (relative to population – these are Recall-like scales: `tpr` = Recall, `1-fpr` = Inverse Recall).

The ROC plot (Fig. 1) allows us to compare classifiers (models and/or parameterizations) and choose the one that is closest to (0,1) and furtherest from `tpr=fpr` in some sense. These conditions for choosing the optimal parameterization or model are not identical. The most common condition is to minimize the area under the curve (AUC), which for a single parameterization of a model is defined by a single point and the segments connecting it to (0,0) and (1,1). A particular cost model and/or accuracy measure defines an isocost gradient, which for a skew and cost insensitive model will be `c=1`, and hence another common approach is to choose a tangent point on the highest isocost line that touches the curve. The simple condition of choosing the point on the curve nearest the optimum point (0,1) is not commonly used, but this distance to (0,1) is given by $\sqrt{[(-\text{fpr})^2 + (1-\text{tpr})^2]}$, and minimizing this amounts to minimizing the sum of squared normalized error, $\text{fpr}^2+\text{fnr}^2$. The area under the simple trapezoid defined by the model is:

```
AUC     = (tpr-fpr+1)/2
        = (tpr+tnr)/2
        = 1 − (fpr+fnr)/2          (9)
```

For the cost and skew insensitive case, with `c=1`, maximizing AUC is thus equivalent to maximizing `tpr-fpr` or minimizing a sum of (absolute) normalized error `fpr+fnr`. The chance line corresponds to `tpr-fpr=0`, and parallel isocost lines for `c=1` have the form `tpr-fpr=k`. The highest isocost line also maximizes `tpr-fpr` and AUC so that these two approaches are equivalent. Minimizing a sum of squared normalized error, $\text{fpr}^2+\text{fnr}^2$, corresponds to a Euclidean distance minimization heuristic that is equivalent only under appropriate constraints, e.g. `fpr=fnr`, or equivalently, Bias=Prevalence, noting that all cells are non-negative by construction.

Another ROC measure studied by Flach [5] is Weighted Relative Accuracy, defined to subtract off the component of the True Positive score that is attributable to chance and rescale to the range ±1. Maximizing WRacc is equivalent to maximizing AUC or `tpr-fpr` = 2·AUC−1, given `c` is constant. Thus WRAcc is an unbiased accuracy measure, and the skew-insensitive form of WRAcc, with `c=1`, is `tpr-fpr`.

## 2.4 DeltaP, Informedness and Markedness

Powers [4] also derived an unbiased accuracy measure to avoid the bias of Recall, Precision and Accuracy due to population Prevalence and label bias. The Bookmaker algorithm costs wins and losses in the same way a fair bookmaker would set prices based on the odds. Powers then defines the concept of Informedness which represents the 'edge' a punter has in making his bet, as evidenced and quantified by his winnings. Fair pricing based on correct odds should be zero sum – that is, guessing will leave you with nothing in the long run, whilst a punter with certain knowledge will win every time. Informedness is the probability that a punter is making an informed bet and is explained in terms of the proportion of the time the edge works out versus ends up being pure guesswork. Powers defined 'Bookmaker Informedness' for the general, K-label, case, but we do not have space here to deal with the general case and present a simplified dichotomous formulation of Powers Informedness, as well as the complementary concept of Markedness.

**Definition 1**

*Informedness quantifies how informed a predictor is for the specified condition, and specifies the probability that a prediction is informed in relation to the condition (versus chance).*

**Definition 2**

*Markedness quantifies how marked a condition is for the specified predictor, and specifies the probability that a condition is marked by the predictor (versus chance).*

These definitions are aligned with the psychological and linguistic uses of the terms condition and marker. The condition represents the experimental outcome we are trying to determine by indirect means. A marker or predictor (cf. biomarker or neuromarker) represents the indicator we are using to determine the outcome. There is no implication of causality – that is something we will address later. However there are two possible directions of implication we will address now. Detection of the predictor may reliably predict the outcome, with or without the occurrence of a specific outcome condition reliably triggering the predictor.

For the dichotomous case we have

```
Informedness  = Recall + Inverse Recall − 1
              = tpr-fpr = 1-fnr-fpr                (10)
Markedness    = Precision + Inverse Precision − 1
              = tpa-fna = 1-fpa-fna                (11)
```

We noted above that maximizing AUC or the unbiased WRAcc measure effectively maximized `tpr-fpr` and indeed WRAcc reduced to this in the skew independent case. This is not surprising given both Powers and Flach set out to produce an unbiased measure, and the linear definition of Informedness will define a unique linear form. Note that while Informedness is a deep measure of how consistently the Predictor predicts the Outcome by combining surface measures about what proportion of Outcomes are correctly predicted, Markedness is a deep measure of how consistently the Outcome has the Predictor as a Marker by combining surface measures about what proportion of Predictions are correct.

In the Psychology literature, Markedness is known as DeltaP and is empirically a good predictor of human associative judgements – that is it seems we develop associative relationships between a predictor and an outcome when DeltaP is high, and this is true even when multiple predictors are in competition [6]. Perruchet and Peereman [7], in the context of experiments on information use in syllable processing, note that [6] identifies DeltaP as the normative measure of contingency, but propose a complementary, backward, additional measure of strength of association, DeltaP' aka Informedness. Perruchet and Peeremant also note the analog of DeltaP to regression coefficient, and that the Geometric Mean of the two measures is a dichotomous form of the Pearson correlation coefficient, the Matthews' Correlation Coefficient, which is appropriate unless a continuous scale is being measured dichotomously in which case a Tetrachoric Correlation estimate would be appropriate [8].

## 2.5 Causality, Correlation and Regression

In a linear regression of two variables, we seek to predict one variable, y, as a linear combination of the other, x, finding a line of best fit in the sense of minimizing the sum of squared error (in y). The equation of fit has the form

$$y = y_0 + r_x \cdot x \quad \text{where}$$
$$r_X = [n\sum x \cdot y - \sum x \cdot \sum y]/[n\sum x^2 - \sum x \cdot \sum x] \quad (12)$$

Substituting in counts from the contingency table, for the regression of predicting **+R** (1) versus **−R** (0) given **+P** (1) versus **−P** (0), we obtain this gradient of best fit (minimizing the error in the real values **R**): $r_P$ = DeltaP = Markedness. Conversely, we can find the regression coefficient for predicting **P** from **R** (minimizing the error in the predictions **P**): $r_R$ = DeltaP' = Informedness.

Finally we see that the Matthews correlation, a contingency matrix method of calculating the Pearson product-moment correlation coefficient, ρ, is defined by: $r_G$ = ±√[Informedness·Markedness].

Given the regressions find the same line of best fit, these gradients should be reciprocal, defining a perfect Correlation of 1. However, both Informedness and Markedness are probabilities with an upper bound of 1, so perfect correlation requires perfect regression. The squared correlation is a coefficient of proportionality indicating the proportion of the variance in R that is explained by P, and is traditionally also interpreted as a probability. We can now interpret it either as the joint probability that P informs R and R marks P, given that the two directions of predictability are independent, or as the probability that the variance is (causally) explained reciprocally. The sign of the Correlation will be the same as the sign of Informedness and Markedness and indicates whether a correct or perverse usage of the information has been made.

Psychologists traditionally explain DeltaP in terms of causal prediction, but it is important to note that the direction of stronger prediction is not necessarily the direction of causality, and the fallacy of abductive reasoning is that the truth of A → B does not in general have any bearing on the truth of B → A.

## 2.6 Effect of Bias & Prev on Recall & Precision

We now recast the Bookmaker Informedness and Markedness equations to show Recall and Precision as subject (13-14), in order to explore the effect of Bias and Prevalence on Recall and Precision, as well as clarify the relationship of Bookmaker and Markedness to these ubiquitous and iniquitous measures.

```
Recall      = Bookmaker (1−Prevalence) + Bias
Bookmaker   = (Recall-Bias) / (1−Prevalence)       (13)
Precision   = Markedness (1-Bias) + Prevalence
Markedness  = (Precision−Prev) / (1-Bias)          (14)
```

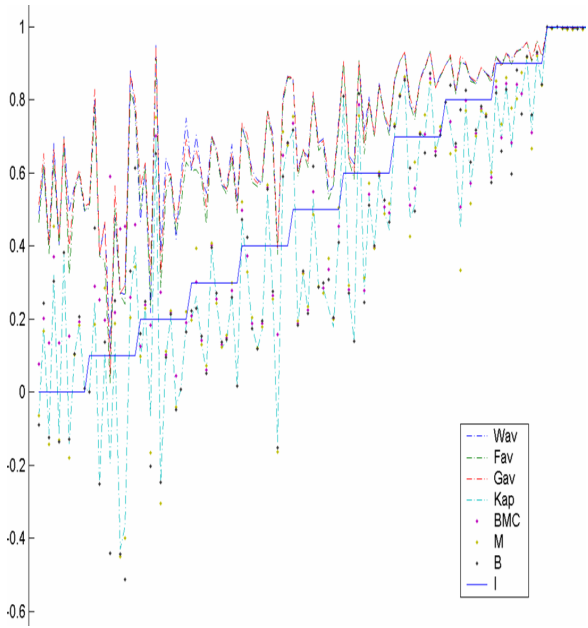

**Figure 2.** Accuracy of traditional measures.
110 Monte Carlo simulations with 11 stepped expected Informedness levels (probability of correct decision versus random binomial decision and random margins) with calculated Informedness, Markedness and Correlation versus Kappa and the biased traditional measures Rank Weighted Average (Wav), Geometric Mean (Gav) and F-factor (Fav).

Bookmaker and Markedness are unbiased estimators of above chance performance (relative to respectively the predicting conditions or the predicted markers). Equations (23-24) clearly show the nature of the bias introduced by both Label Bias and Class Prevalence. If operating at chance level, both Bookmaker and Markedness will be zero, and Recall, Precision, and derivatives such as the F-measure, will merely reflect the biases. Note that increasing Bias or decreasing Prevalence increases Recall and decreases Precision, for a constant level of unbiased performance. We can more specifically see that the regression coefficient for the prediction of Recall from Prevalence is –Bookmaker and from Bias is +1, and similarly the regression coefficient for the prediction of Precision from Bias is –Markedness and from Prevalence is +1.

Alternately, Informedness can be viewed (13) as a renormalization of Recall after subtracting off the chance level of Recall, Bias. Markedness (14) can be seen as a renormalization of Precision after subtracting off the chance level of Precision, Prevalence. Cohen's Kappa measure [9,10] commonly used in assessor agreement evaluation was similarly defined as a renormalization of Accuracy after subtracting off the expected Accuracy as estimated by the dot product of the Biases and Prevalences. All three renormalized measures are invariant in the sense that they are properties of the contingency tables that remain unchanged when we flip to the Inverse problem (interchange positive and negative for both conditions and predictions).

Although Kappa does renormalize a debiased estimate of Accuracy, and is thus much more meaningful than Recall, Precision, Accuracy, and their biased derivatives, it is intrinsically non-linear, doesn't account for error well, and retains an influence of bias, so that there does not seem that there is any situation when Kappa would be preferable to Correlation as a standard independent measure of agreement [11,12,8]. Bookmaker Informedness, Markedness and Correlation, as their geometric mean, reflect the discriminant, $dp$, of relative contingency normalized according to different functions of the marginal Biases and Prevalences, and reflect probabilities relative to the corresponding marginal cases. However, Kappa scales in a way that reflects the actual error without taking into account expected error due to chance, and in effect it is really just scaling the actual mean error: Kappa is $dp/[dp+\text{mean}(fp,fn)]$ which for small error approximates as $1-\text{mean}(fp,fn)/dp$.

The relatively good fit of Kappa to Correlation and Informedness is illustrated in Fig. 2, along with the poor fit of the Rank Weighted Average and the Geometric and Harmonic (F-factor) means.

## 3 CONCLUSIONS

The system of relationships we have discovered is amazingly elegant and easy to teach as meaningful probabilities: Informedness is simply Recall + Inverse Recall – 1, whilst Markedness is simply Precision + Inverse Precision – 1. Correlation is their geometric mean. The non-linear Cohen Kappa is the next best alternative and approximates reasonably in the dichotomous case, especially when Bias tracks Prevalence and Informedness and Markedness collapse to Correlation as Recall and Precision collapse to Rand Accuracy.